# Finding links and initiators: a graph-reconstruction problem

Heikki Mannila[*]        Evimaria Terzi[†]

.

## Abstract

Consider a 0–1 observation matrix $M$, where rows correspond to entities and columns correspond to signals; a value of 1 (or 0) in cell $(i, j)$ of $M$ indicates that signal $j$ has been observed (or not observed) in entity $i$. Given such a matrix we study the problem of inferring the underlying directed links between entities (rows) and finding which entries in the matrix are initiators.

We formally define this problem and propose an MCMC framework for estimating the links and the initiators given the matrix of observations $M$. We also show how this framework can be extended to incorporate a temporal aspect; instead of considering a single observation matrix $M$ we consider a sequence of observation matrices $M_1, \ldots, M_t$ over time.

We show the connection between our problem and several problems studied in the field of social-network analysis. We apply our method to paleontological and ecological data and show that our algorithms work well in practice and give reasonable results.

## 1 Introduction

Analyzing 0-1 matrices is one of the main themes in data mining. Techniques such as clustering or mixture modelling, matrix decomposition techniques such as PCA, ICA, and NMR, and Bayesian all aim to give an answer to the informal question: "Where does the matrix come from?" These approaches aim at describing a probabilistic generative model that describes the observed matrix well.

In this paper we consider yet another way of answering the question "Where does a 0–1 matrix $M$ come from?" In our model, the matrix $M$ of size $n \times m$ is considered to arise from *initiators*, certain few entries that are initially 1. The initiators *propagate* their 1's by following the links of a directed influence graph $G$ (represented by an $n \times n$ adjacency matrix). We denote the initiator matrix of size $n \times m$ by $N$ and we use $G$ (of size $n \times n$) to refer both to the directed graph between the rows of $M$ and as well as its adjacency matrix. Then, we believe that the structure of $N$ and $G$ can tell how a matrix $M$ has been created. As an example, consider the following data matrix $M$:

$$
\begin{array}{c|cccccc}
r_1 & 1 & 1 & 1 & 0 & 0 & 0 \\
r_2 & 1 & 1 & 1 & 1 & 1 & 1 \\
r_3 & 0 & 0 & 0 & 1 & 1 & 1
\end{array}
$$

We can describe this matrix by considering the $3 \times 6$ initiator matrix $N$ and $3 \times 3$ adjacency matrix $G$:

$$
\begin{array}{c|cccccc}
 & & & & & & \\
r_1 & 1 & 1 & 1 & 0 & 0 & 0 \\
r_2 & 0 & 0 & 0 & 0 & 0 & 0 \\
r_3 & 0 & 0 & 0 & 1 & 1 & 1
\end{array}
\qquad
\begin{array}{c|ccc}
 & r_1 & r_2 & r_3 \\
r_1 & 0 & 1 & 0 \\
r_2 & 0 & 0 & 0 \\
r_3 & 0 & 1 & 0
\end{array}
$$

The directed graph $G$ tells that row $r_1$ influences row $r_2$, and that also row $r_3$ influences row $r_2$.

The structure of $G$ and $N$ of course depends on the assumed propagation model. For the purposes of this paper we use a propagation model that is similar in spirit to the ones used in [8, 11, 15, 17]. Informally, matrices $G$ and $N$ specify the probability for each entry of $M$ being 1, as follows: the positions that already have value 1 in $N$ will have value 1 in $M$ (almost) certainly. Otherwise, a value of 1 in entry $N(a, j)$ propagates to entries $M(b, j)$ with probability $p$, where $p$ is proportional to $\alpha^{-L}$ and $L$ is the length of shortest path from $b$ to $a$ in $G$. If there are multiple entries $N(a, j) = 1$ such that $a$ and $b$ are connected in $G$, the probabilities are combined in the natural way. Thus, given $G$ and $N$ the propagation model gives a probability of each entry in $M$ being 1 or 0. From these probabilities we can compute the likelihood $\Pr(M|G, N)$.

---

[*]Helsinki Institute of Information Technology, University of Helsinki and Helsinki University of Technology, Finland. Email: `heikki.mannila@tkk.fi`

[†]IBM ARC, San Jose, CA 95120. Email: `eterzi@us.ibm.com`

In our case though, $G$ and $N$ are unknown and the task is to estimate them given a matrix of observations $M$. As a data-analysis task this setting can seem to be quite underspecified: after all, we start from an $n \times m$ matrix $M$, and as output we obtain an $n \times m$ matrix $N$ *and* $n \times n$ adjacency matrix $G$. Thus, there are more bits in the output than there are in the input. For example, we could always take $N = M$ and let $G$ be empty. We adopt a Bayesian approach to handle this issue. We prefer solutions where the number $x$ of 1s in $N$ is small, so we put a prior for $N$ that is proportional to $c^{-x}$ for some constant $c > 1$; a similar prior on the number of edges in $G$ implements our preference for graphs with a small number of edges.

Our task is then the following: given an observed 0–1 matrix $M$, sample graph-initiator pairs $(G, N)$ from the posterior distribution

$$\Pr(G, N|M) \propto \Pr(G)\Pr(N)\Pr(M|G, N).$$

From such samples, we can then compute the average strength that an edge is present in $G$ or that a particular entry is present in $N$.

We show that this approach yields intuitive results on real ecological and paleontological data sets. While the method converges relatively slowly, it is still usable for data sets with hundreds of rows and columns.

**Related work:** Here we give a brief summary of related work. The problem of inferring a network structure given a set of observations is not new; there has been lots of research for finding graphical models (see for example [4, 10] and references therein) and Bayesian networks in particular (see [5, 12] and references therein). Network discovery problems have also been studied a lot in biological applications where the focus is to to find metabolic networks or gene regulatory networks given a set of observations, see [14] for a 37-page bibliography on these themes. Although our problem is related to the problems studied in the above papers, this relationship remains at a fairly high level. For example, we are not aware of any graph reconstruction problem that assumes our information-propagation model. Moreover, the aspect of identifying the initiators as well as the underlying link structure that we bring here is new.

To a certain extent the work of [1] and [16] is also related to ours. The authors in these papers try to identify influential individuals and initiators by looking at blog postings ([1]) and purchase data ([16]). However, the aspect of estimation of links between the individuals is missing there.

Finally, our problem is related to the problem of identifying nodes in a social network to achieve maximum diffusion of information as studied in [8, 11] or the opposite problem of identifying nodes in a network to maximally contain the spread of an epidemic studied for example in [2, 7, 13]. The main difference between these problems and ours is that we do not assume any given network as part of the input. A more detailed discussion about the connection of our problem to the maximum information-propagation problems studied in the past is given in Section 6.

**Roadmap:** The rest of this paper is organized as follows: in Section 2 we describe the propagation model we assume that given a graph determines the transition of 1-entries from the initiators to the receivers. In Section 3 we describe our algorithm for sampling pairs of graphs and initiators from the right posterior distribution and in Section 4 we show how to extend this algorithm in the presence of more than one instances of the observation matrices. Some indicative experiments on real data are given in Section 5. In Section 6 we discuss the problem using a more general viewpoint and point out connections to existing work. We conclude in Section 7.

## 2 The propagation model

Consider $n \times m$ 0–1 matrix $M$ such that every row of the matrix corresponds to an entity and every column correspond to a signal. The entries of the matrix take values $\{0, 1\}$ and $M(i, j) = 1$ ($M(i, j) = 0$) is interpreted signal $j$ has been observed (not observed) at entity $j$. We call matrix $M$ the *observation matrix*. Since our matrices contain only 0's and 1's, we also give to them a set interpretation.

The underlying assumption is that the rows of $M$ (the entities) are organized in an underlying (directed) graph $G$. The interactions of the nodes in this graph determine the formation of the final matrix $M$. Additionally, let $N$ be an $n \times m$ 0–1 matrix that gives information about the initiators of the propagation. That is, if $N(i,j) = 1$, then the $i$-th entity has initiated the propagation of signal $j$. We call matrix $N$ the *initiators matrix*.

Assume a (known) directed network $G$ and initiators matrix $N$. We consider the following information-propagation model: a signal $u$ is observed at node $i$, that is not an initiator for $u$, if $u$ is transferred to $i$ from one of the initiators of $u$ via a path in $G$.

The longer the path from an initiator $j$ to $i$ the less influence node $j$ has in the appearance of $u$ in node $i$. We use function $b(j,i,G)$ to denote the influence that node $j$ has to node $i$ in graph $G$. If we use $d_G(j,i)$ to denote the length of the shortest path from $j$ to $i$ in the graph $G$, then we have

$$b(j,i,G) \quad = \quad \alpha^{d_G(j,i)}, \tag{1}$$

where $\alpha \in [0,1]$. The intuition behind the form of Equation (1) is that the influence a node had on any other node in the graph decays exponentially with their graph distance. We call this propagation model the *shortest path* (SP) model.

Next we define the likelihood of the observed data $M$ given $G$ and $N$. Assuming that all entries in $M$ are independent we have that

$$\Pr(M|G,N) \quad = \quad \prod_{i=1}^{n} \prod_{u=1}^{m} \Pr(M(i,u)|G,N).$$

Each individual term $\Pr(M(i,u)|G,N)$ is easy to define. First recall that each entry $M(i,u)$ can take values 0 or 1. The case $M(i,u) = 0$ occurs when no 1 in the $u$ column of $N$ propagates to row $i$ and $N(i,u) = 0$. That is,

$$\begin{aligned} \Pr(M(i,u) = 0|G,N) = \\ &= \quad (1 - N(i,u)) \prod_{j \neq i} (1 - N(j,u)b(j,i,G)) \\ &= \quad \prod_{j=1}^{n} \left( 1 - N(j,u)\alpha^{d_G(j,i)} \right). \end{aligned}$$

For the case $M(i,u) = 1$ we have

$$\Pr(M(i,u) = 1|G,N) \quad = \quad 1 - \Pr(M(i,u) = 1|G,N)$$

**Value of the parameter $\alpha$:** For a given graph $G$, initiator matrix $N$ and observation matrix $M$, the correct value of the parameter $\alpha$ is easily inferred to be the one that maximizes the probability of the data $M$ given $G$ and $N$.

# 3 Sampling graphs and initiators

## 3.1 Basics

Our input is the observations matrix $M$ of size $n \times m$, and the goal is to sample graphs and initiators from the posterior probability distribution $\Pr(G,N|M)$. Note that we are interested in *directed* graphs and therefore every time we refer to a graph we will imply that it is directed, unless specified otherwise. Using Bayes rule we get that $\Pr(G,N|M)$ is equivalent to the following expression:

$$\begin{aligned} \Pr(G,N|M) \quad &= \quad \frac{\Pr(M|G,N)\Pr(G,N)}{\Pr(\mathrm{M})} \tag{2} \\ &\propto \quad \Pr(M|G,N)\Pr(G,N) \tag{3} \\ &= \quad \Pr(M|G,N)\Pr(G)\Pr(N). \tag{4} \end{aligned}$$

Step (2) in the above derivation is due to Bayes rule, Step (3) is due to the fact that $\Pr(M)$ is constant and finally, Step 4 comes from the assumption that $G$ and $N$ are independent. Alternatively, we could assume that the position of a node in a graph affects it as being an initiator or not; in this case instead of $\Pr(G,N) = \Pr(G)\Pr(N)$ we would have $\Pr(G,N) = \Pr(G)\Pr(N|G)$. Despite this being a reasonable assumption, for simplicity we assume independence in this paper.

Therefore, given a constant normalizing factor $Z$, our goal is to actually sample from the distribution

$$\pi(G,N|M) \quad = \quad \frac{1}{Z}\Pr(M|G,N)\Pr(N)\Pr(G). \tag{5}$$

The term $\Pr(M|G,N)$ was already discussed in the previous section. The other two terms of Equation (5), $\Pr(G)$ and $\Pr(N)$ encode our prior beliefs about the form of the graph and the initiator matrix. In principle we want both $G$ and $N$ to be as condensed as possible. That is, we want to penalize for graphs with large number of edges, and for initiator matrices with lots of initiators initiating the same product. We encode these beliefs by penalizing for any additional 1-entry in $G$ and in $N$. Thus we set, $\Pr(G) \propto e^{-c_1|E|}$ and $\Pr(N) \propto e^{-c_2|N|}$. Parameters $c_1$ and $c_2$ are constants, $|N|$ is the number of 1-entries in matrix $N$ and $|E|$ is the number of edges in $G$.

## 3.2 MCMC algorithm

In this section we show how to sample graph-initiator pairs from the posterior probability distribution $\pi$. We start by describing a naive way of sampling graph-initiator pairs uniformly at random by means of a random walk on a Markov chain $\mathcal{C}$. The state space of $\mathcal{C}$ are all the possible graph-initiator pairs. Then we show how to modify this chain to allow for getting samples from the right posterior probability $\pi$.

Consider the following *naive* algorithm for sampling the state space of graph-initiator pairs. Start with a graph-initiator pair $(G_0, N_0)$ and perform a *local moves* to other graph-initiator pairs. When many local moves have been performed, consider the resulting pair $(G, N)$ in the sample. We call this algorithm the `Naive` algorithm, and in order to fully specify it we only need to additionally define what we mean by "local moves".

A local move from graph-initiator pair $(G, N)$ to the pair $(G', N')$ can be defined by a cell $(i, j)$ $(i \neq j)$ in $G$ or a cell $(k, \ell)$ in $N$. The new pair $(G', N')$ is formed by updating $G$ and $N$ so that $G'(i,j) = 0$ $(= 1)$ if $G(i,j) = 1$ $(= 0)$, and $N'(k,\ell) = 1$ $(= 0)$ if $N(k,\ell) = 0$ $(= 1)$. More specifically, given $(G, N)$ a `LocalMove` routine first uniformly picks $G$ or $N$ to modify and then performs one of the changes described above.

Formally, a local move is a step on a Markov chain $\mathcal{C} = \{\mathcal{S}, \mathcal{T}\}$, where the state space $\mathcal{S}$ consists of the set of all pairs of graphs and initiator matrices; $\mathcal{T}$ is the set of transitions defined by local moves. In other words, the set $\mathcal{T}$ contains all pairs $\big((G,N),(G',N')\big)$ such that it is possible to obtain $(G',N')$ from $(G, N)$ (or vice versa) via local moves.

Notice that the Markov chain $\mathcal{C}$ (a) is *connected* and (b) all the states in $\mathcal{S}$ have the same degree. Therefore chain $\mathcal{C}$ converges to a uniform stationary distribution. As a result the `Naive` sampling algorithm also samples graph-initiator pairs uniformly at random.

However, we want to sample graphs and initiators from the posterior probability distribution $\pi$ (Equation (5)). This is done by using the *Metropolis-Hastings* algorithm. In every state $(G, N)$ the transition to state $(G', N')$ suggested by a random local move is accepted with probability

$$\min\big\{1, \frac{\pi(G', N'|M)}{\pi(G, N|M)}\big\}.$$

Algorithm 1 gives the pseudocode for the `Metropolis-Hastings` algorithm.

**Quantities to monitor:** As noted above, our goal is not to find a single pair $(G, N)$ that would have a high likelihood. Rather, we want to estimate the posterior probabilities that a link exists between two nodes, or that a particular entry is an initiator. To do this, we collect (after a sufficient burn-in period) statistics on how many times each entry in the matrices $G$ and $N$ is present. We denote the average values of $G$ by $\widehat{G}$ and the averages of the initiator matrix by $\widehat{N}$. We report these quantities in our experiments.

**Starting state and running times:** In principle a run of the `Metropolis-Hastings` algorithm can start from any arbitrary state and converge to the desired stationary probability distribution. However, this may take exponentially many steps. In our experiments we use as a starting state the graph-initiator pair $(G_0, N_0)$ where $N_0 = M$ and $G_0$ is the empty graph.

**Algorithm 1** `Metropolis-Hastings`

---
1: **Input:** An initial state $(G_0, N_0)$ and a number of random walk steps $s$
2: **Output:** A pair of graph and initiator $(G, N)$
3: $(G, N) \leftarrow (G', N')$
4: **while** $s > 0$ **do**
5:     $(G', N') \leftarrow$ `LocalMove`$(G, N)$
6:     $(G, N) \leftarrow (G', N')$ with probability $\min\left\{1, \frac{\pi(G', N')}{\pi(G, N)}\right\}$
7:     $s \leftarrow s - 1$
8: **end while**
9: return $(G, N)$

---

Given the current state $(G, N)$, every iteration of the `Metropolis-Hastings` algorithm requires a computation of the posterior probability distribution for the proposed state $(G', N')$. If the proposed state $(G', N')$ is such that $G' = G$ then the computation of the posterior probability $\pi$ requires $O(n^2m)$ time. If on the other hand $G' \neq G$ and $N' = N$, then the computation of the posterior probability requires additionally a computation of all-pairs-shortest paths on the newly formed graph $G'$ and therefore it requires $O(n^2m + n^2 \log n)$ time.

## 4 Incorporating temporal information

Instead of a single observation matrix $M$, assume an ordered (by time) sequence of observation matrices $M_1, \ldots, M_T$. That is, $M_i$ corresponds to the observation matrix at some timestamp $i$. The assumption is that the set of 1-entries in matrix $M_t$ is a superset of the 1-entries of matrix $M_{t-1}$ for every $t \in \{1, \ldots, T\}$, i.e., for every $1 \leq t < T$ we have $M_t \subseteq M_{t+1}$. For example, once a signal is observed at some entity, then this is a recorded fact and cannot be cancelled.

As before, given observation matrices $M_1, \ldots, M_T$ our goal is to sample graph-initiator pairs from the posterior distribution of graphs and initiators given the observation matrices. Note that in this case every time instance $t$ is associated with its own initiator matrix $N_t$. Therefore, in this case we are dealing with graph-initiator pairs but rather with pairs of graphs and sequences of initiators $(G, \widetilde{N})$, where $\widetilde{N} = \{N_1, \ldots, N_T\}$. If $N_t(i, u) = 1$, then it means that signal $u$ has been initiated by $i$ at time $t$. We use $\widetilde{N}$ to denote the sequence of initiators $N_1, \ldots, N_T$ and $\widetilde{M}$ to denote the sequence of observation matrices $M_1, \ldots, M_T$. Moreover, we use $\widetilde{N}_t$ and $\widetilde{M}_t$ to denote the sequence of the $t$ first elements of the initiators and observations sequences respectively. Therefore $\widetilde{N}_T = \widetilde{N}$ and $\widetilde{M}_T = \widetilde{M}$.

Using again Bayes rule we get that the posterior probability $\pi_T$ from which we want to sample is

$$
\begin{aligned}
\pi_T(G, \widetilde{N} | M) &= \Pr(G, \widetilde{N} | \widetilde{M}) \\
&= \frac{\Pr(\widetilde{M} | G, \widetilde{N}) \Pr(G, \widetilde{N})}{\Pr(\widetilde{M})} \\
&\propto \Pr(\widetilde{M} | G, \widetilde{N}) \Pr(G, \widetilde{N}) \\
&= \Pr(\widetilde{M} | G, \widetilde{N}) \Pr(G) \Pr(\widetilde{N}).
\end{aligned} \tag{6}
$$

We have

$$
\begin{aligned}
\Pr(\widetilde{N}) &= \Pr(N_1, \ldots, N_T) \\
&= \Pr(N_1) \times \ldots \times \Pr(N_T).
\end{aligned}
$$

Using now the definition of conditional probability, the probability $\Pr(\widetilde{M} | G, \widetilde{N})$ can be easily written

as follows:

$$\Pr(\widetilde{M}_T | G, \widetilde{N}_T)$$
$$= \Pr(M_T | G, \widetilde{M}_{T-1}, \widetilde{N}_T) \times \Pr(\widetilde{M}_{T-1} | G, \widetilde{N}_T) \tag{7}$$
$$= \Pr(M_T | G, M_{T-1}, N_T) \times \Pr(\widetilde{M}_{T-1} | G, \widetilde{N}_{T-1}) \tag{8}$$
$$= \dots$$
$$= \Pr(M_T | G, M_{T-1}, N_T) \times \Pr(M_{T-1} | G, M_{T-2}, N_{T-1})$$
$$\times \dots \times \Pr(M_1 | G, N_1).$$

In the above equation, line (7) is by the definition of conditional probability. In line (8) we substitute $\Pr(M_T | G, \widetilde{M}_{T-1}, \widetilde{N}_T)$ with $\Pr(M_T | G, M_{T-1}, N_T)$, because of the assumption that $M_t \subseteq M_{t+1}$. Moreover we substitute $\Pr(\widetilde{M}_{T-1} | G, \widetilde{N}_T)$ with $\Pr(\widetilde{M}_{T-1} | G, \widetilde{N}_{T-1})$ since instances $M_1, \dots, M_{T-1}$ can only depend on initiators $N_1, \dots, N_{T-1}$ and cannot depend on the initiators of time $T$.

As before, for an observation matrix $M_t$ we have that

$$\Pr(M_t(i, u) = 0 | G, M_{t-1}, N_t) =$$
$$= \prod_{j=1}^{n} \Big( 1 - M_{t-1}(j, u) b(j, i, G) - N_k(j, u) b(j, i, G)$$
$$+ M_{t-1}(j, u) N_k(j, u) b(j, i, G) \Big)$$
$$= \prod_{j=1}^{n} \Big( 1 - \big( M_{t-1}(j, u) + N_k(j, u)$$
$$- M_{t-1}(j, u) N_k(j, u) \big) \alpha^{d_G(j, i)} \Big).$$

That is, the probability that $M_t(i, u) = 0$ is equal to the the probability that $i$ does not get $u$ from any of the nodes that had it at some previous point in time neither did it get it from any of the nodes that initiated $u$ at time $t$. Naturally, the probability that $M_t(i, u) = 1$ is

$$\Pr(M_t(i, u) = 1 | G, M_{t-1}, N_{t-1}) =$$
$$= 1 - \Pr(M_t(i, u) = 0 | G, M_{t-1}, N_{t-1}).$$

The `Metropolis-Hastings` algorithm described in the previous section can be applied for the sampling of pairs of graphs and initiator sequences $(G, \widetilde{N})$ from the posterior probability distribution $\pi'$. There are three differences though. First, the state space of the Markov chain will consist of the pairs of all possible graphs and initiator sequences. A local move from state $(G, \widetilde{N})$ to $(G' \widetilde{N}')$ would randomly pick between changing an entry $(i, j)$ in $G$ and flip it from 1 to 0 (or from 0 to 1) or one entry $(k, \ell)$ in each one of the initiator matrices in the sequence $\widetilde{N}$. Finally, given current state $(G, \widetilde{N})$ a proposed state $(G', \widetilde{N}')$ is accepted with probability $\min\{1, \pi(G', \widetilde{N}' | \widetilde{M}) / \pi(G, \widetilde{N} | \widetilde{M})\}$.

**Quantities to monitor:** As in the case of a single observation matrix here we also collect (after a sufficient burn-in period) statistics on how many times each entry in the matrices $G$ and $N_t$ (for $1 \leq t \leq T$) is present. We denote the average values of $G$ by $\widehat{G}$ and the averages of the initiator matrix at time $t$ by $\widehat{N}_t$. We are also interested in the overall average initator matrix $\widehat{N}_{\text{all}}$ which is the average over all initiator matrices $N_t$. That is, $\widehat{N}_{\text{all}} = 1/T \sum_{t=1}^{T} \widehat{N}_t$. We report these quantities in our experiments.

**Starting state and running times:** Obviously the state space of the Markov chain constructed for this extended version of the problem is huge. Therefore, despite it being connected, we cannot hope that the sampling algorithm will converge in polynomial time. We start the Markov chain from the state $(G_0, \widetilde{N}^0)$, where $G_0$ is the empty graph and for every $N_t \in \widetilde{N}^0$ it holds that $N_t = M_t \setminus M_{t-1}$.

Given the current state $(G, \widetilde{N})$, every iteration of the `Metropolis-Hastings` for the case of multiple instances posterior probability for the proposed state $(G', \widetilde{N}')$. If the proposed state $(G', \widetilde{N}')$ is such that $G' = G$ then the computation of the posterior probability $\pi'$ requires $O(Tn^2m)$ time. If on the other hand $G' \neq G$ and $\widetilde{N}' = \widetilde{N}$, then the computation of the posterior probability requires additionally a computation of all-pairs-shortest paths on the newly formed graph $G'$ and therefore it requires $O(T(n^2m + n^2 \log n))$ time.

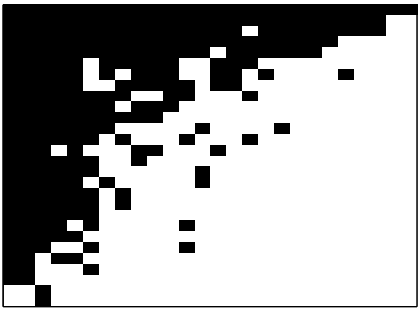

Figure 1: The original **Rocky Mountain** dataset consisting of 28 sites and 26 species.

## 5 Experiments

In this section we show some indicative experimental results we obtained by applying our methods to ecological and paleontological datasets. The 0–1 matrices in these domains represent the presence/absence of species to sites. For each such presence/absence matrix we experiment with, the rows correspond to the sites and the columns correspond to the species. Our goal when analyzing these datasets is to find connection between sites with respect to, say, the migration of species from one site to another. For example, the existence of an edge between site $i$ and site $j$, with direction from $i$ to $j$ would mean that there is high confidence that species appeared first to site $i$ and migrated to site $j$. Initiators are also important; if a site is estimated to be an initiator for a specific species, then this is the site in which the species first appeared and then it migrated elsewhere.

The goal of our experiments is to illustrate that our method can provide useful insights in the analysis of such datasets, that cannot be obtained by other standard data-analysis methods.

For all our experiments we set $c_1 = 2$ and $c_2 = 9$, since we believe that the cost of an initiator is larger than the cost of a single connection in a graph. For the `Metropolis-Hastings` algorithm, we split the sampling process into two stages: the burn-in period and the sampling period. All the experiments are run for $2 \times 10^5$ steps. We use the first $10^5$ steps as the burn-in period and the rest $10^5$ for actual sampling. Every $10^4$ samples obtained during the actual sampling steps we report the instance of the average graph constructed by these $10^4$ samples. In that way, we have 10 different estimates of the average graph, $\widehat{G}_1, \ldots, \widehat{G}_{10}$ ordered based on the iteration at which they were obtained. Similarly we get ten different estimates of the average initiator matrix $\widehat{N}_1, \ldots, \widehat{N}_{10}$. We refer to the last sample $\widehat{G}_{10}$ (or $\widehat{N}_{10}$) as the average graph (or the average initiator matrix) and we denote them by $\widehat{G}$ ($\widehat{N}$ respectively).

For a given dataset, we run the sampling algorithm for values of $\alpha$ in $\{0.1, \ldots, 0.9\}$ and we pick the one for which the sampled graph-initiator pairs converge to the largest value for the posterior probability.

### 5.1 Experiments with ecological data

The ecological datasets used for the experiments are available by AICS Research Inc, University Park, New Mexico, and The Field Museum, Chicago. The datasets are available online[1] and they have been used for a wide range of ecological studies [3, 6, 18].

We focus our attention on the results we obtained by applying our method to a single such dataset; the **Rocky Mountain** dataset. The dataset shows the presence/absence of Boreal and boreo-cordilleran species of mammals in the Southern Rocky Mountains, and has been used as a dataset of reference for many ecological studies, see for example [3]. The dataset itself is rendered in Figure 1.[2]

Now consider this dataset and let the data analysis task be the inference of directed relationships between sites with respect to migration of species between them. Lets first consider dealing with this task by using an off-the-shelf clustering algorithm. The idea would be to first cluster the rows of the dataset into, say, $k$ groups, and then infer very strong connections among sites in the same group and weaker (or no) connections among sites in different groups. The clustering of the sites of the **Rocky Mountain** dataset into $k = 2$ clusters and an assignment of weight 1 to the connections between sites in the same group, and weight 0 to the connections between the sites results in inferring the (crude) graph structure appearing in Figure 2(a). For the clustering we used standard k-means algorithm and measured the distance between points using the Hamming distance. Notice in the plot the ordering of the sites is the same as their ordering in the original dataset in Figure 1. Therefore, the clustering algorithm infers strong connections between the species-rich sites and strong connections between the species-poor sites, and no connections between them.

---

[1] http://www.aics-research.com/nestedness/
[2] Throughout we use black to plot the 1-entries of a 0–1 matrix and white to plot the 0-entries.

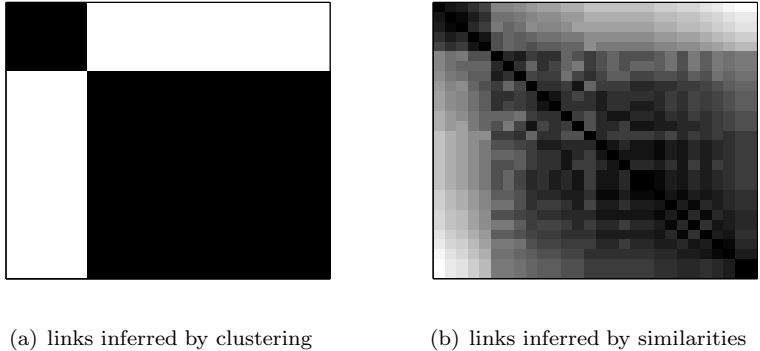

(a) links inferred by clustering

(b) links inferred by similarities

Figure 2: **Rocky Mountain** data - Figure 2(a): connection between sites by strongly connecting sites in the same cluster and not connecting sites in different clusters. Figure 2(b): pairwise similarity values between sites.

An alternative way of inferring connections between sites would be by defining a similarity function between sites and then conclude that high similarity values between two sites implies strong migration connection between them. The inferred connections between sites when using as similarity the inverse of the Hamming distance between the sites is shown in Figure 2(b).

Although the pairwise connections between sites extracted by using clustering or similarity computations can give some insight on the connections between sites, it is obvious that the connection strengths inferred by these computations are rather crude. Moreover, neither of these two methods can give us some insight as of which sites have acted as the initiator for certain species.

Figure 3 shows the results we obtained when analysing the same dataset using our method; in Figures 3(a) and 3(c) we plot the average graph $\widehat{G}$ and the average initiator matrix $\widehat{N}$ obtained by the `Metropolis-Hastings` algorithm for the **Rocky Mountain** dataset. The parameter $\alpha$ was set to $\alpha = 0.9$, since this was the value of $\alpha$ that was steadily converging to graph-initiator pairs having the highest posterior probability $\pi$. Again, the ordering of the sites in the rows of $\widehat{N}$ and in the rows and columns of $\widehat{G}$ retain their original ordering in the dataset shown in Figure 1. Recall that the $\widehat{G}$'s are directed and thus the corresponding adjacency matrices are not expected to be symmetric. The elements in the main diagonal of the adjacency matrix of $\widehat{G}$ are all 1's since they represent connections of nodes with themselves.

The average graph-initiator pair $(\widehat{G}, \widehat{N})$ is rather surprising. The average graph $\widehat{G}$, reveals much different set of connections between sites than those inferred by doing simple clustering or similarity computations. In fact, $\widehat{G}$ reveals that for this value of $\alpha$, there are only *local* interactions between the sites. That is, the interactions between sites with similar concentration of species is stronger than the interactions between sites with different species concentration.

In terms of species migration Figure 3(a) suggests that species migrate from sites with smaller number of species to sites with larger number of species. Given the structure of the dataset, this also suggests that every site acts as an initiator for the species that none of its subsets contains. This is verified by the rendering of the $\widehat{N}$, average matrix of initiators shown in Figure 3(c).

In order to test the reliability of the obtained results, we also plot the correlation coefficients between the estimations of $\widehat{G}$ and $\widehat{N}$ at different steps of the sampling phase of the `Metropolis-Hastings` algorithm. Figures 3(b) and 3(d) show the pairwise correlation coefficients between the average graphs

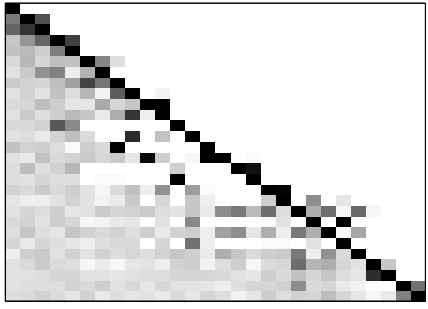

(a) $\widehat{G}$

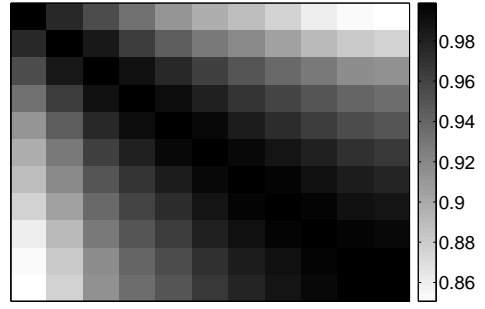

(b) cor. coef. of $\widehat{G}_i$'s

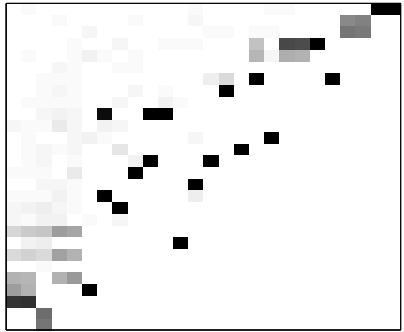

(c) $\widehat{N}$

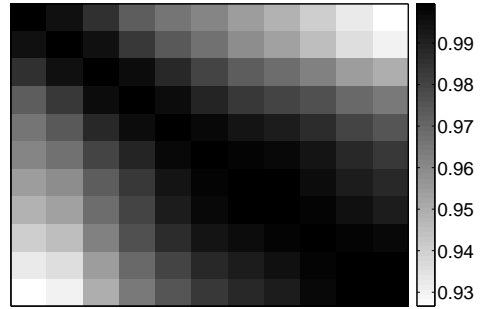

(d) cor. coef. of $\widehat{N}_i$'s

Figure 3: Rocky Mountain data - `Metropolis-Hastings` results for $\alpha = 0.9$. Figure 3(a): average graph $\widehat{G}$. Figure 3(b): pairwise correlation coefficients between the average graphs $\widehat{G}_1, \ldots, \widehat{G}_{10}$ obtained every $10^4$ samples in the the sampling stage of the `Metropolis-Hastings` algorithm. Figure 3(c): and average initiator matrix $\widehat{N}$. Figure 3(d): pairwise correlation coefficients between the average initiator matrices $\widehat{N}_1, \ldots, \widehat{N}_{10}$ obtained every $10^4$ samples in the the sampling stage of the `Metropolis-Hastings` algorithm.

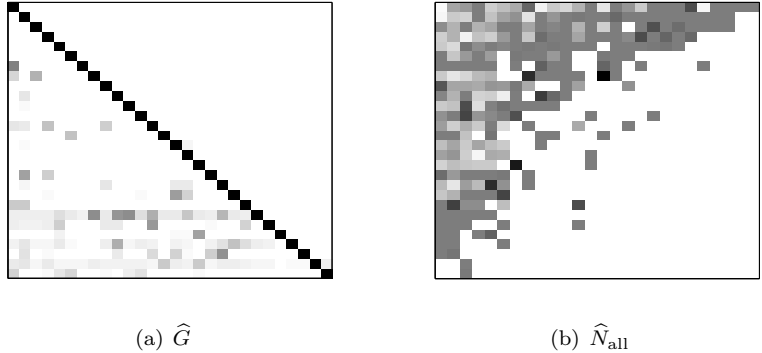

(a) $\widehat{G}$          (b) $\widehat{N}_{\text{all}}$

Figure 4: **Rocky Mountain** data with $T = 3$ artificially generated instances $M_1, M_2, M_3$ - $M_3$ corresponds to the original data. Results of the `Metropolis-Hastings` algorithm for $\alpha = 0.9$. Figure 4(a): average graph $\widehat{G}$. Figure 4(b): average initiator matrix $\widehat{N}_{\text{all}}$.

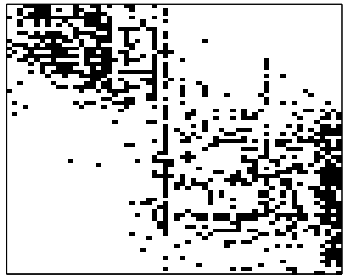

Figure 5: The original paleontological dataset consisting of 60 sites and 70 species.

$\widehat{G}_i$'s and the average initiator matrices $\widehat{N}_i$'s. These average graphs and initiators are obtained during the sampling phase of the `Metropolis-Hastings` algorithm in a fashion described in the beginning of the experimental section. Note that all instances of $\widehat{G}_i$'s and $\widehat{N}_i$'s are highly correlated with each other since the lowest value of the correlation coefficient observed is 0.86 and 0.93 respectively, while most of the values are very close to 1. This verifies that our sampling algorithm converges to the desired posterior probability distribution.

Naturally, one can ask the question of how different these results would be if instead of a single observation matrix $M$, we had several instances $M_1, \ldots, M_T$ that would show how the distribution of species to sites evolves over time. Given that such data is not available for the **Rocky Mountain** dataset, we artificially generate it as follows: we set $M_T = M$ and for $t = 1, \ldots, T - 1$ we generate instance $M_t$ by converting every 1-entry in $M_{t+1}$ to 0 with probability $1/T$. In that way $M_t \subseteq M_{t+1}$ as we assumed in Section 4. In Figure 4 we show the average graph $\widehat{G}$ and the average initiator matrix $\widehat{N}_{\text{all}}$ we obtained using $T = 3$ instances of the **Rocky Mountain** dataset generated as described above. The value of $\alpha$ was again set to 0.9. The average graph $\widehat{G}$ shown in Figure 4(a) shows an obvious similarity with the average graph $\widehat{G}$ shown in Figure 3(a); in both graphs the relatively strong links appear to start from sites with small species concentration towards sites with larger species concentration. The average initiator matrix $\widehat{N}_{\text{all}}$ in Figure 4(b) is much denser than the corresponding $\widehat{N}$ obtained from the single instance dataset in Figure 3(c). However, in this case as well it is obvious there there is a band of high values across the diagonal of the matrix that spans the matrix from southwest to northeast.

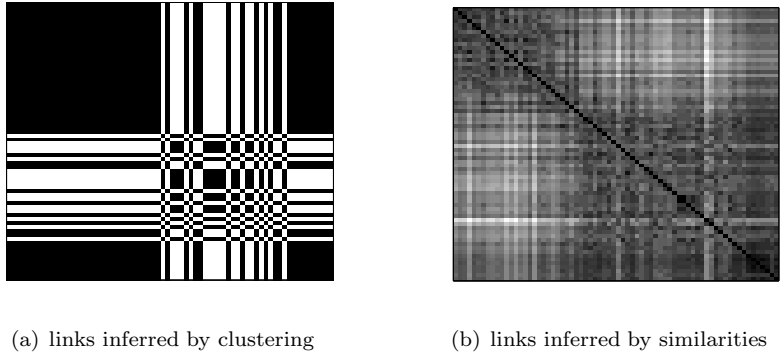

(a) links inferred by clustering       (b) links inferred by similarities

Figure 6: Paleontological data - Figure 6(a): connection between sites by strongly connecting sites in the same cluster and not connecting sites in different clusters. Figure 6(b): pairwise similarity values between sites.

## 5.2 Experiments with paleontological data

We present here a similar set of experiments for a paleontological dataset. The dataset contains information of fossil genera found in palaeontological sites in Europe [9].[3] The original dataset shown in Figure 5 exhibits some kind of block structure, since by visual inspection one can identify to distinct blocks in the rendered matrix. Figures 6(a) and 6(b) show the graph structures among sites that are inferred using $k$-means clustering ($k = 2$) or similarity computations among the rows of the input matrix. As before, the estimated graphs appear to be rather crude, and give no information about the initiators of species.

Figures 7(a) and 7(c) show the $\widehat{G}$ and $\widehat{N}$ estimated by the `Metropolis-Hastings` sampling algorithm. For this experiment we used the same burn-in and sampling scheme as the one described in the beginning of the experimental section. For this dataset we show the results for $\alpha = 0.4$, since this was the value of $\alpha$ for which the `Metropolis-Hastings` algorithm steadily converged to samples of graph-initiator pairs with the highest likelihood $\pi$. Both the estimated $\widehat{G}$ and $\widehat{N}$ show that the connections between the sites in these case are not so strong; matrix $\widehat{G}$ contains lots of 0-entries, while there are quite many initiators in $\widehat{N}$, when compared to the original matrix $N$. This is partly due to the fact that there are not really sites in which very similar sets of species appear; thus for every edge introduced in the graph we have to pay for a considerable number of false transfers of 1's. At the same time the optimal value of $\alpha$ in this case being 0.4, has as a result to encourage significant number of initiators; since for relatively small values of $\alpha$ transfers of 1's are not likely to succeed.

Figures 7(b) and 7(d) show the the correlation coefficients between all pairs of $\widehat{G}_i$'s and $\widehat{N}_i$'s obtained during the sampling period of the run of the `Metropolis-Hastings` algorithm. Notice that in this case the instances of of $\widehat{G}_i$'s and $\widehat{N}_i$'s are not as correlated as in the case of the **Rocky Mountain** dataset. However the still exhibit high correlation since the smallest values obtained in these matrices are 0.6 and 0.86 respectively, while most of the values are very close to 1, verifying our assumption that the algorithm indeed converges reasonably fast.

## 6 Extensions and connections to other problems

**Other information propagation models:** In the previous sections we have described our approach for estimating graphs and initiators given an arbitrary 0–1 observation matrix $M$. We have studied this problem for a specific information propagation model, namely the shortest path propagation model (SP) (see Section 2). We have focused our discussion to SP because it is simple and intuitive. However, our framework for estimating graphs and initiators can be used with any information propagation model, like for example the models studied in [11, 17].

Consider an arbitrary (not necessarily deterministic) information propagation model $\mathcal{P}$ that given a graph $G$ and an initiator matrix $N$ produces a matrix $M_p = \mathcal{P}(G, N)$. That is, the 1's from the initiating

---

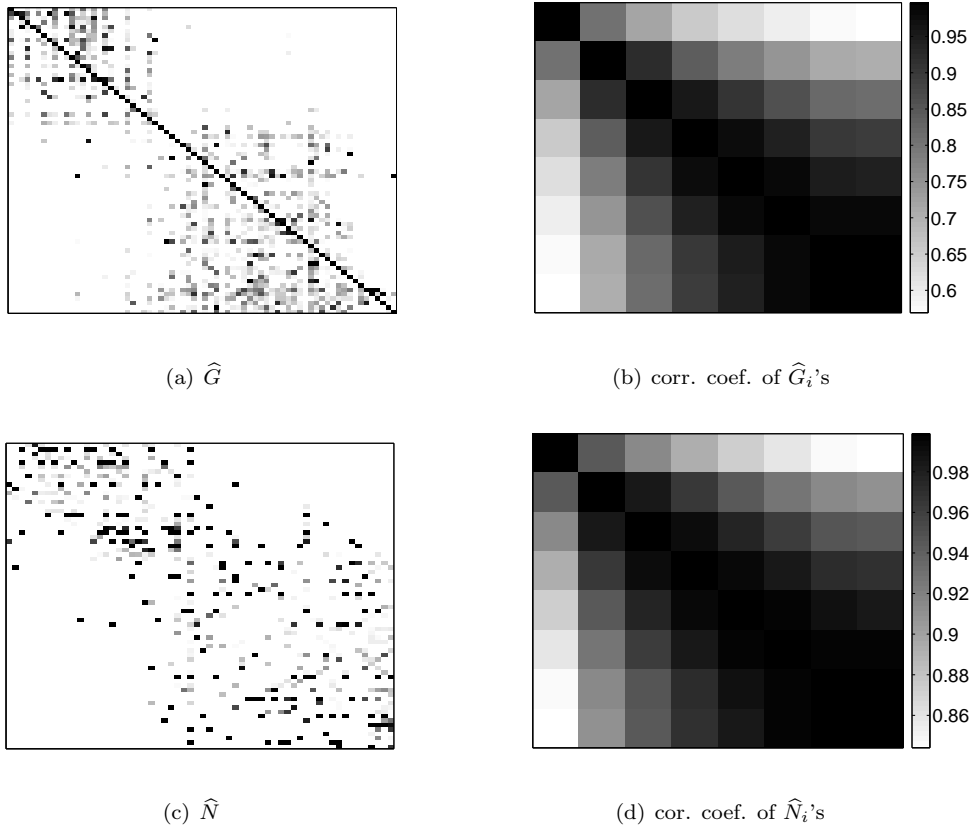

(a) $\widehat{G}$

(b) corr. coef. of $\widehat{G}_i$'s

(c) $\widehat{N}$

(d) cor. coef. of $\widehat{N}_i$'s

Figure 7: Paleontological dataset - `Metropolis-Hastings` results for $\alpha = 0.4$. Figure 7(a): average graph $\widehat{G}$. Figure 7(b): pairwise correlation coefficients between the average graphs $\widehat{G}_1, \ldots, \widehat{G}_{10}$ obtained every $10^4$ samples in the the sampling stage of the `Metropolis-Hastings` algorithm. Figure 7(c): and average initiator matrix $\widehat{N}$. Figure 7(d): pairwise correlation coefficients between the average initiator matrices $\widehat{N}_1, \ldots, \widehat{N}_{10}$ obtained every $10^4$ samples in the the sampling stage of the `Metropolis-Hastings` algorithm.

rows of $N$ are propagated to other rows via the edges of the graph $G$ and according to the propagation rules determined by the propagation model $\mathcal{P}$. In that way matrix $M_p = \mathcal{P}(G, N)$ is formed.

Now assume that the problem that we considered before. Namely, given a 0–1 observation matrix $M$ and propagation model $\mathcal{P}$ estimate the structure (existence and strength of connections) of a directed graph $G$ and the initiators $N$. In other words, we again ask for a sample of graphs and initiators from the posterior probability distribution

$$\Pr(G, N|M) \quad \propto \quad \Pr(M|G, N)\Pr(G)\Pr(N).$$

Thus the task is the same as before. The only term in the above formula that depends on the propagation model is the term $\Pr(M|G, N)$. However, since $G$ and $N$ are known, matrix $M_p = \mathcal{P}(G, N)$ is also known. Therefore for some constant $c$ we can substitute $\Pr(M|G, N)$ with

$$\Pr(M|G, N) \quad \propto \quad e^{-c|M - M_p|} = e^{-c|M - \mathcal{P}(G, N)|}. \tag{9}$$

Note that in the above equation $|M - M_p|$ refers to the sum of absolute values of the entries of matrix $M - M_p$. Therefore, Equation (9) penalizes for graphs and initiators that generate matrices $M_p$ that are considerably different from the observed matrix $M$. Therefore, under any propagation model $\mathcal{P}$, Equation (9) can be used to compute the probability of the observed matrix $M$ given $G$ and $N$. With this, we can deploy our sampling algorithms under any propagation model.

Notice that in general, the information propagation models do not have to be deterministic and consequently neither does the matrix $M_p$. For non-deterministic propagation models $M_p = E\big[\mathcal{P}(G, N)\big]$. That is, in these cases $M_p$ is the expectation of the constructed matrix, where the expectation is taken over the random choices of the propagation model.

**Connections to other problems:** In the preceding discussion we have assumed that only the observation matrix $M$ is given and the goal was to estimate the strengths of directed links between the rows of $M$, as well as the beliefs for certain rows being initiators for signals that correspond to certain columns. Here, we discuss some optimization problems that arise in the case where apart from the observation matrix $M$, also the graph $G$ is given as part of the input. The natural question to ask in this case is formalized in the following problem definition.

**Problem 1** (Minimum Initiation problem). *Given an $n \times m$ 0–1 observation matrix $M$, graph $G$, propagation model $\mathcal{P}$ and integer $k$, find an initiator matrix $N$ with with at most $k$ 1-entries so that $|M - \mathcal{P}(G, N)|$ is minimized.*

We can very easily observe, that this problem is already hard for many well-known information propagation models, like for example the *linear threshold* (LT) and the *independent cascade* (IC) model described in [11]. Here we are not giving a detailed description of these two propagation models, we refer to [11] for this. For the rest of the discussion we can treat them as a black-box, bearing in mind that they are non-deterministic. We state the hardness result of the Minimum Initiation problem in the observation below and we discuss it immediately after.

**Observation 1.** *For $\mathcal{P}$ being either the linear threshold (LT) or the independent cascade (IC) models, the Minimum Initiation problem is NP-hard.*

The above observation is an immediate consequence of [11]. A careful observation of the results in [11], show that the Minimum Initiation problem is NP-hard even for the case where the observation matrix $M$ of size $n \times 1$, all entries of $M$ have value 1 and $|M - \mathcal{P}(G, N)|$ is required to be 0. The following inapproximability result is an immediate consequence of this observation.

**Corollary 1.** *When $\mathcal{P}$ is either the linear threshold (LT) or the independent cascade (IC) models, there is no polynomial-time approximation algorithm $\mathcal{A}$ for the Minimum Initiation problem, such that the solution of $\mathcal{A}$ is within a factor $\alpha > 1$ from the optimal solution to the Minimum Initiation problem, unless P=NP.*

# 7 Conclusions

We have introduced a model for finding the links and initiators that could be responsible for the observed data matrix. We described a simple propagation model for generating data from links and initiators, and gave an MCMC algorithm for sampling from the right posterior distribution of graphs and initiators. We also studied some decision versions of the problem and showed their connections to related work from social-network analysis.

The empirical results show that the MCMC approach is – perhaps surprisingly – able to find meaningful structure in this setting. While there is no unique answer, the runs converge to certain average values of the graph-initiator pairs.

Obviously there are lots of open questions in the approach. An interesting question is selection among the different propagation models. The MCMC method converges fairly slowly, and therefore it would be of interest to see whether the individual iterations could be made faster. Ecological interpretation of the results is among the themes that we are currently pursuing actively.

# References

[1] N. Agarwal, H. Liu, L. Tang, and P. S. Yu. Identifying the influential bloggers in a community. In *WSDM*, pages 207–218, 2008.

[2] J. Aspnes, K. L. Chang, and A. Yampolskiy. Inoculation strategies for victims of viruses and the sum-of-squares partition problem. *J. Comput. Syst. Sci.*, 72(6):1077–1093, 2006.

[3] W. Atmar and B. Patterson. On the measure of order and disorder in ecological communities on archipelagos. *Oecologia*, 96:539–547, 1993.

[4] W. L. Buntine. A guide to the literature on learning probabilistic networks from data. *IEEE Trans. Knowl. Data Eng.*, 8(2):195–210, 1996.

[5] D. M. Chickering and C. Meek. Finding optimal bayesian networks. In *UAI*, pages 94–102, 2002.

[6] A. Cutler. Nested biotas and biological conservation: metrics, mechanisms, and meaning of nestedness. *Landscape and Urban Planning*, 28:73–82, 1994.

[7] Z. Dezso and A.-L. Barabasi. Halting viruses in scale-free networks. *Phys. Rev. E*, 66, 2002.

[8] P. Domingos and M. Richardson. Mining the network value of customers. In *KDD*, pages 57–66, 2001.

[9] M. Fortelius. Neogene of the old world database of fossil mammals (NOW). http://www.helsinki.fi/science/now/, 2008.

[10] D. Geiger and D. Heckerman. Probabilistic relevance relations. *IEEE Transactions on Systems, Man, and Cybernetics, Part A*, 28(1):17–25, 1998.

[11] D. Kempe, J. M. Kleinberg, and É. Tardos. Maximizing the spread of influence through a social network. In *Proceedings of the ACM SIGKDD International Conference on Knowledge Discovery and Data Mining (KDD)*, pages 137–146, 2003.

[12] M. Koivisto. Advances in exact bayesian structure discovery in bayesian networks. In *UAI*, 2006.

[13] J. Leskovec, A. Krause, C. Guestrin, C. Faloutsos, J. VanBriesen, and N. S. Glance. Cost-effective outbreak detection in networks. In *KDD*, pages 420–429, 2007.

[14] F. Markowetz. A bibliography on learning causal networks of gene interactions, 2008.

[15] S. Morris. Contagion. *Review of Economic Studies*, 67, 2000.

[16] P. Rusmevichientong, S. Zhu, and D. Selinger. Identifying early buyers from purchase data. In *KDD*, pages 671–677, 2004.

[17] Y. Wang, D. Chakrabarti, C. Wang, and C. Faloutsos. Epidemic spreading in real networks: An eigenvalue viewpoint. In *22nd Symposium on Reliable Distributed Computing (SRDS)*, 2003.

[18] D. Wright, B. D. Patterson, G. M. Mikkelson, A. Cutler, and W. Atmar. A comparative analysis of nested subset patterns of species composition. *Oecologia*, 113:1–20, 1998.